\documentclass[letterpaper, 10 pt, conference]{ieeeconf}  % Comment this line out if you need a4paper
\IEEEoverridecommandlockouts                              % This command is only needed if 
\usepackage{xcolor}
\usepackage{tabularx}
\usepackage{cite,graphicx,amsmath,amssymb,array,rotating}
\usepackage{hyperref}
\title{\LARGE \bf
	Towards Task-Specific Modular Gripper Fingers: Automatic Production of Fingertip Mechanics 
}

\author{Johannes Ringwald, Samuel Schneider, Lingyun Chen, Dennis Knobbe, Lars Johannsmeier,\\ Abdalla Swikir and Sami Haddadin% <-this % stops a space
	\thanks{This work was funded by the German Research Foundation (DFG, Deutsche Forschungsgemeinschaft) as part of Germany’s Excellence Strategy – EXC 2050/1 – Project ID 390696704 – Cluster of Excellence “Centre for Tactile Internet with Human-in-the-Loop” (CeTI) of Technische Universität Dresden. The authors gratefully acknowledge the funding of the Lighthouse Initiative KI.FABRIK Bayern by StMWi Bayern (KI.FABRIK Bayern Phase 1: Aufbau Infrastruktur and KI.Fabrik Bayern Forschungs- und Entwicklungsprojekt, grant no. DIK0249).
We also gratefully acknowledge the funding of the Lighthouse Initiative Geriatronics by LongLeif GaPa gGmbH (Project Y).The authors are with the Chair of Robotics and Systems Intelligence and Munich Institute of Robotics and Machine Intelligence, Technical University Munich (TUM), D-80797 Munich, Germany {\tt\small e-mail: \{firstname.surname\}@tum.de}}%
	\thanks{$^{\star}$ Please note that S. Haddadin has a potential conflict of interest as a shareholder of Franka Emika GmbH.}
	\thanks{$^{\star}$ This work has been submitted to the IEEE for possible publication. Copyright may be transferred without notice, after which this version may no longer be accessible.}
}%

\begin{document}

	\maketitle
	\thispagestyle{empty}
	\pagestyle{empty}
	
	%%%%%%%%%%%%%%%%%%%%%%%%%%%%%%%%%%%%%%%%%%%%%%%%%%%%%%%%%%%%%%%%%%%%%%%%%%%%%%%%
	% Text
	\begin{abstract}
The number of sequential tasks a single gripper can perform is significantly limited by its design.
In many cases, changing the gripper fingers is required to successfully conduct multiple consecutive tasks. For this reason, several robotic tool change systems have been introduced that allow an automatic changing of the entire end-effector. However, many situations require only the modification or the change of the fingertip, making the exchange of the entire gripper uneconomic. In this paper, we introduce a paradigm for automatic task-specific fingertip production. The setup used in the proposed framework consists of a production and task execution unit, containing a robotic manipulator, and two 3D printers - autonomously producing the gripper fingers. It also consists of a second manipulator that uses a quick-exchange mechanism to pick up the printed fingertips and evaluates gripping performance. The setup is experimentally validated by conducting automatic production of three different fingertips and executing grasp-stability tests as well as multiple pick- and insertion tasks, with and without position offsets - using these fingertips. The proposed paradigm, indeed, goes beyond fingertip production and serves as a foundation for a fully automatic fingertip design, production and application pipeline - potentially improving manufacturing flexibility and representing a new production paradigm: tactile 3D manufacturing. 
\end{abstract}
	\section{Introduction}
% Finger adaption
%\newtext{Opt-1: The task and environmental conditions like position and orientation of manipulation objects can be controlled well in an industrial production setup, while design, control and a successful application of complex multi DoF hands are still subject of basic research. Also, the performance of soft-grippers depend on the considered scenario. Potential robustness problems can arise if objects with sharp edges must be grasped with high grasping forces.Accordingly, most industrial assembly tasks are still performed by two- or three-finger parallel grippers.}
Most conditions of manipulation tasks in industrial production setups can be precisely controlled, like initial positions and orientations of manipulation objects which should be processed. Despite these controllable conditions, performing tasks with multi-degree of freedom (DoF) hands is still a complex assignment and a subject of basic research. Moreover, the performance of soft grippers depends on the scenario considered. For example,  robustness problems can arise if objects with sharp edges must be grasped with high grasping forces. 
Accordingly, most industrial assembly tasks are still performed by two- or three-finger parallel grippers.
To enable robust grasping and manipulation of defined objects, the fingers of these grippers must be manually designed, produced, tested and iterated until reaching a satisfactory performance. This adaptation process is not only very laborious and time-consuming, but also requires a high level of design experience in this field \cite{ha2020fit2form}. Therefore, the number of assembly objects and manipulation scenarios that can be handled is limited by the time a designer needs to adapt a finger to a specific application.
This makes the assembly line setup in-flexible, since any changes, like tool-changing, finger adaption etc. are very time consuming and therefore costly. Accordingly, assembly or production lines are usually setup for a few product types, which are produced in masses over a longer time period (month/years). All integrated robots conduct only one specific task with one specialized tool or finger design, which increases the required number of robots for a designated assembly scenario significantly. \\
%A flexible production of smaller batch sizes, which can change the type of produced products in a small amount of time (minutes/hours), would require significantly shorter - ideally fully automatized - gripper finger design, production and iteration steps.
A flexible production of smaller batch sizes requires a smaller adaption time of the setup to a new product (minutes/hours). Therefore, a significantly shorter - ideally fully automatized - gripper finger design, production and iteration process would be needed.
%------------------------------------
% Picture: General Pipeline/Approach
\begin{figure}[t!]
	\centerline{\includegraphics[width=8.7cm]{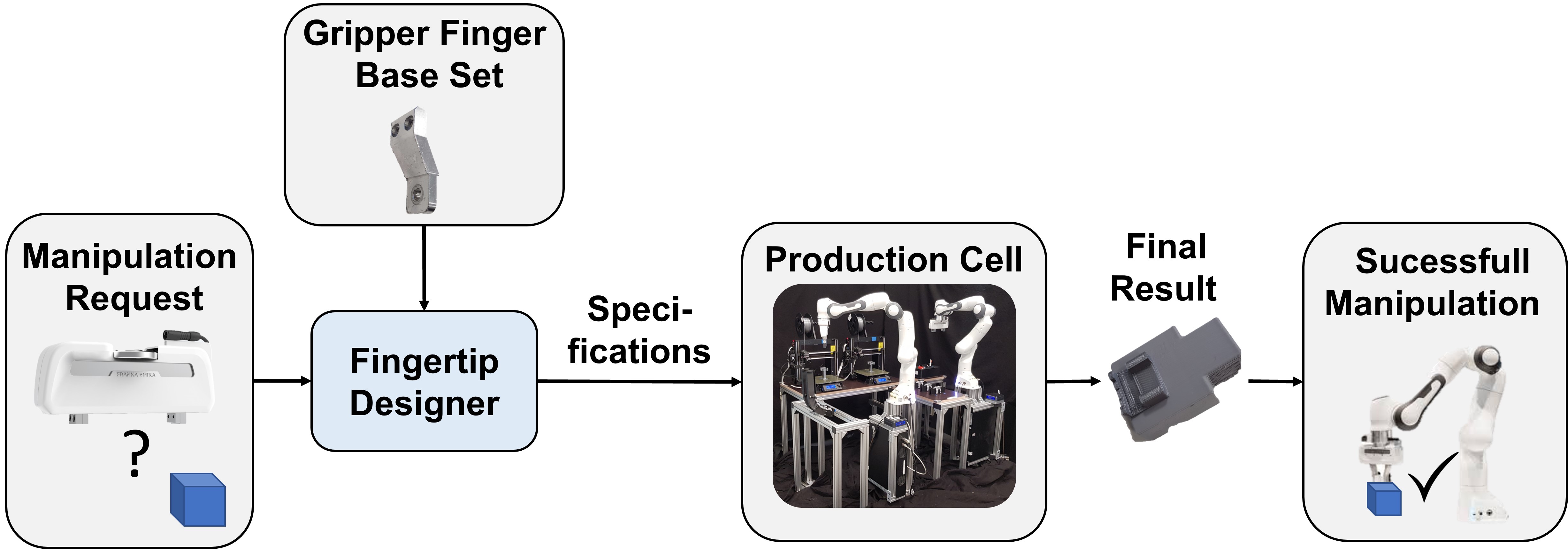}}
	\caption{%
	Automatic fingertip production and task execution based on a given fingertip design and finger base set - using two robot-arms and two 3D printers.
	\vspace{-0.5cm}
    }
	\label{fig:Main-Setup}
\end{figure}
%------------------------------------
%------------------------------------
% Picture: Main-Setup Visualization
\begin{figure*}[ht!]
	\centerline{\includegraphics[width=15cm]{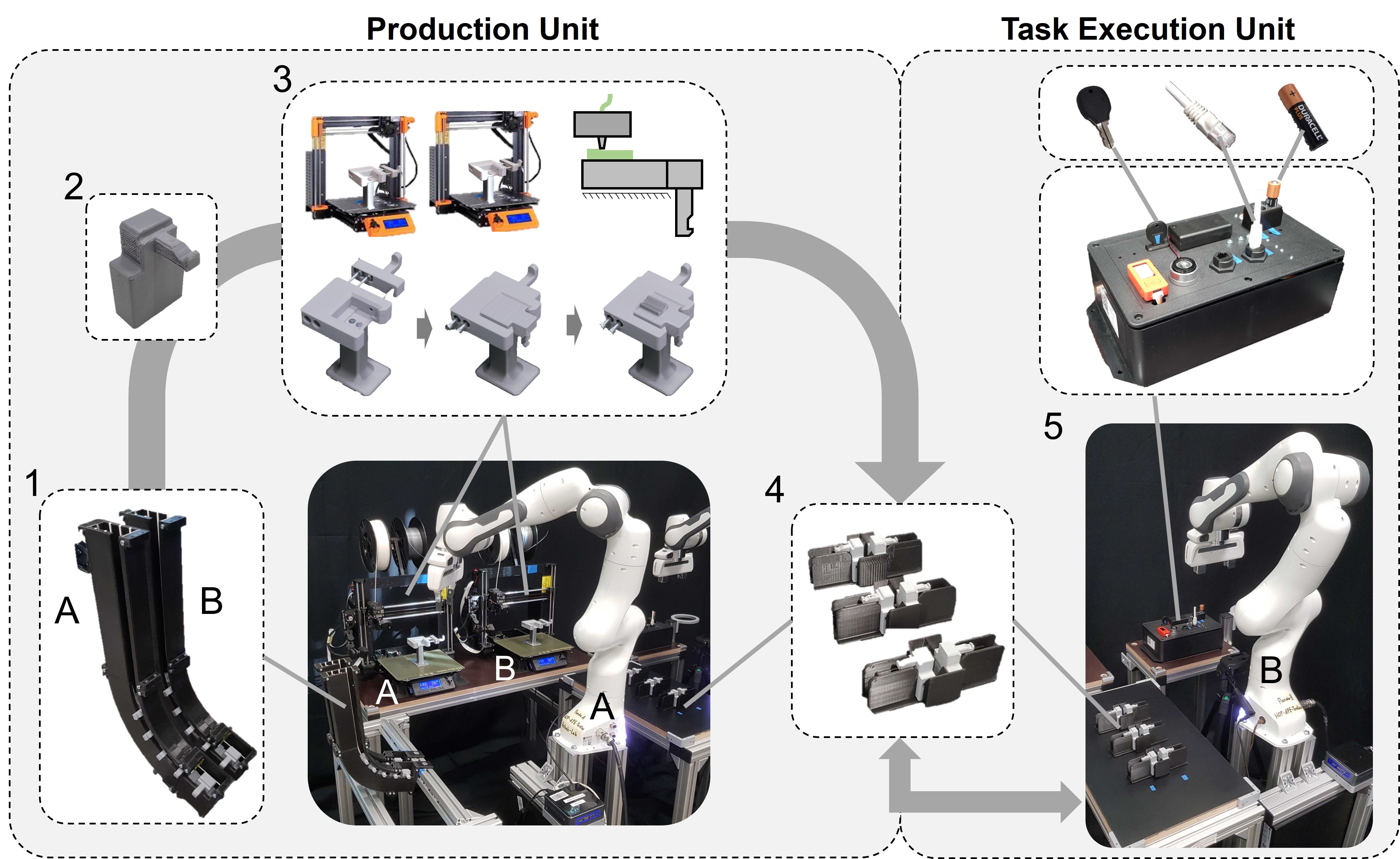}}
	\caption{Presented testbed divided into a production and a task-execution unit.
	% Production Unit
	The production unit conducts the extension of 3D printed standard finger-bases with custom fingertip elements by: 1. extraction of two finger-bases from corresponding finger-base magazines by a robot-arm. 2. Insertion and locking of these finger-bases to finger-holder frames, stationary placed on the build-plate of two 3D printers. 3. Printing of custom fingertips on top of pre-produced finger-bases. 4. The extended finger is picked up by the robot-arm and inserted into the magazines of the quick-finger-exchange system. 5. Using the QFE system, the gripper fingers are picked up by the task execution unit.
	% Task Execution Unit
	To conduct the finger testing tasks an IoT-Box is used, which provides pick and place tasks for three different objects: key, Ethernet-cable, battery.\vspace{-0.3cm}
	}
	\label{fig:Production/Task-Units}
\end{figure*}
%------------------------------------

% Automatic finger adaption
Several automatic approaches for robot finger design adaptation have been introduced to solve the aforementioned problems \cite{brown19993, Pedrazzoli2001, Pham1992, Pham2007, Saravanan2009, chen2018topology, Liu2018, Liu2020optimal}.
One aspect of this adaptation process concerns the finger surface design which majorly affects the ability to robustly perform stable grasps and manipulation tasks.
Different approaches have been introduced to optimize the gripper finger morphology and geometry based on a given set of objects \cite{Honarpardaz2017a, Honarpardaz2017b, Honarpardaz2017c, Honarpardaz2019, ha2020fit2form, Song2018, Balan2003, Velasco1998computer}.
% Robot Tool-Changing systems
Despite all the efforts to automate the gripper finger design, the set of tasks a robot manipulator can perform is still limited to its mounted fingers. 
Therefore, different tool-changing systems for robotic applications have been developed \cite{homberg2015haptic,leitao2019development, ryuh2006automatic, clevy2005micromanipulation, chagouri2021product, ambrosio2014design, brillowski1996sturdy, dhakal2019design, babcinschi2018automatic, Pettinger2019PassiveTC, berenstein2018open, Gyimothy2011ExperimentalEO, friedman2007automated, mckinley2016interchangeable,backus2016adaptive,ren2018heri}.
% Finger changing systems
While most of these systems focus on exchanging the entire end-effector, some of them focus specifically on replacing single fingers \cite{homberg2015haptic, backus2016adaptive, clevy2005micromanipulation}, or modifying the finger configuration \cite{ren2018heri}.
Unfortunately, none of the aforementioned work provides a fully automated process for the gripper finger production and exchange. In other words, the fingers must be replaced manually during an ongoing manipulation process.
% % Picture: General Pipeline/Approach
% \begin{figure}[t!]
% 	\centerline{\includegraphics[width=8.7cm]{Figures/Automatic-Finger-Design-and-Production-Pipeline_v2-02.jpg}}
% 	\caption{%
% 	Automatic fingertip production and task execution based on a given fingertip design and finger base set - using two robot-arms and two 3D printers.
% 	\vspace{-0.5cm}
%     }
% 	\label{fig:Main-Setup}
% \end{figure}
% %-----------------------------------
% % Picture: Main-Setup Visualization
% \begin{figure*}[ht!]
% 	\centerline{\includegraphics[width=15cm]{Figures/Production-and-Task-Execution-Unit_v1-03.jpg}}
% 	\caption{Presented testbed divided into a production and a task-execution unit.
% 	% Production Unit
% 	The production unit conducts the extension of 3D printed standard finger-bases with custom fingertip elements by: 1. extraction of two finger-bases from corresponding finger-base magazines by a robot-arm. 2. Insertion of these finger-bases to finger-holder frames, stationary placed on the build-plate of two 3D printers. 3. Printing of custom fingertips on top of pre-produced finger-bases. 4. The extended finger is picked up by the robot-arm and inserted into the magazines of the quick-finger-exchange system. 5. Using the QFE system, the gripper fingers are picked up by the task execution unit.
% 	% Task Execution Unit
% 	To conduct the finger testing tasks an IoT-Box is used, which provides pick and place tasks for three different objects: key, Ethernet-cable, battery.\vspace{-0.5cm}
% 	}
% 	\label{fig:Production/Task-Units}
% \end{figure*}
% %-----------------------------------
%

% Own approach
Motivated by the above limitations, we introduce a fully automated setup for gripper finger production and exchange. In particular, by integrating a collaborative robot - the Panda arm - with two 3D printers, it was possible to automate the production of a predefined set of fingertips. 
Furthermore, we introduce a passive quick-finger-exchange system that allows the robot to automatically change the used finger-pair during a manipulation task. 
%\oldtext{By combining the automatic fingertip production and the quick-finger-exchange system, our setup increases the number of tasks and applications a single robot can perform.} 
By combining the automatic fingertip production and the quick-finger-exchange system, the time required to adapt to a new assembly task dramatically decreases. E.g., the automatic production of new fingertips using a standard finger base reduces the production time of new gripper fingers significantly (minutes instead of hour). \\
The number of assembly steps a single robot can perform during a new production/assembly task increases, since one robot can select and change its gripper-finger automatically during the entire assembly process.
Accordingly our approach makes it possible to realize an efficient and flexible small batch production/assembly process, while minimizing the required time, labor and infrastructure required for the assembly - like the number of robots needed for the task. Fig.\ref{fig:Main-Setup} illustrates this automatic process. Moreover, the proposed approach will serve as a first step and foundation for a fully automated task-specific finger design, production, and application pipeline,  which will enable a robot to identify its own fingertips and, potentially, provide the base for a novel production paradigm: tactile 3D manufacturing.

% Structure of the paper
The paper is organized as follows. In Sec. \ref{pts}, we describe the production processes as well as the quick-finger-exchange mechanism. Then, in Sec. \ref{exp}, we demonstrate the application and effectiveness of the proposed approach through experiments. Finally, we discuss our results and provide a conclusion in Sec. \ref{cond}.
	\section{Production and Task Execution Unit}\label{pts}
The proposed setup is divided into a production unit and a task execution unit, see Fig.\ref{fig:Production/Task-Units}. 
The production unit performs all the required steps to automatically build a task-specific parallel gripper fingertip design onto a pre-defined finger-base. 
The task execution unit uses these fingers to conduct a defined set of tasks. Those fingers can be automatically changed during the robot operation using a quick-exchange system.
Accordingly, the fingers the production unit provides can be collected to a finger library and used multiple times by the task-execution unit - depending on the currently needed fingertip geometry. This allows the same robots of an assembly line/setup to change their fingers automatically in seconds and adapt to a new product quickly. \\
%(Opt-2) This allows a quick change of the product to be assembled, since the same robots of a single production line can select and change their fingers automatically in seconds.}
Therefor, our specifically designed 3D-printing based production setup has the following advantages compared to a traditional non-automated manually conducted gripper finger production process: (1) The presented setup is able to react to new tasks and required finger geometries quickly since no manual work or other production services  are required. (2) This level of automation safes time consuming and potentially costly labor. (3) Since no regular worker needs to operate the setup, our approach is easily scalable to an arbitrary number of production units without any limitations due to skilled worker shortage. (4) Our setup can serve as a base framework for a fully automatic gripper finger design, production and task execution pipeline.
% Picture: ...
\begin{figure}[ht!]
	\centerline{\includegraphics[width=8.5cm]{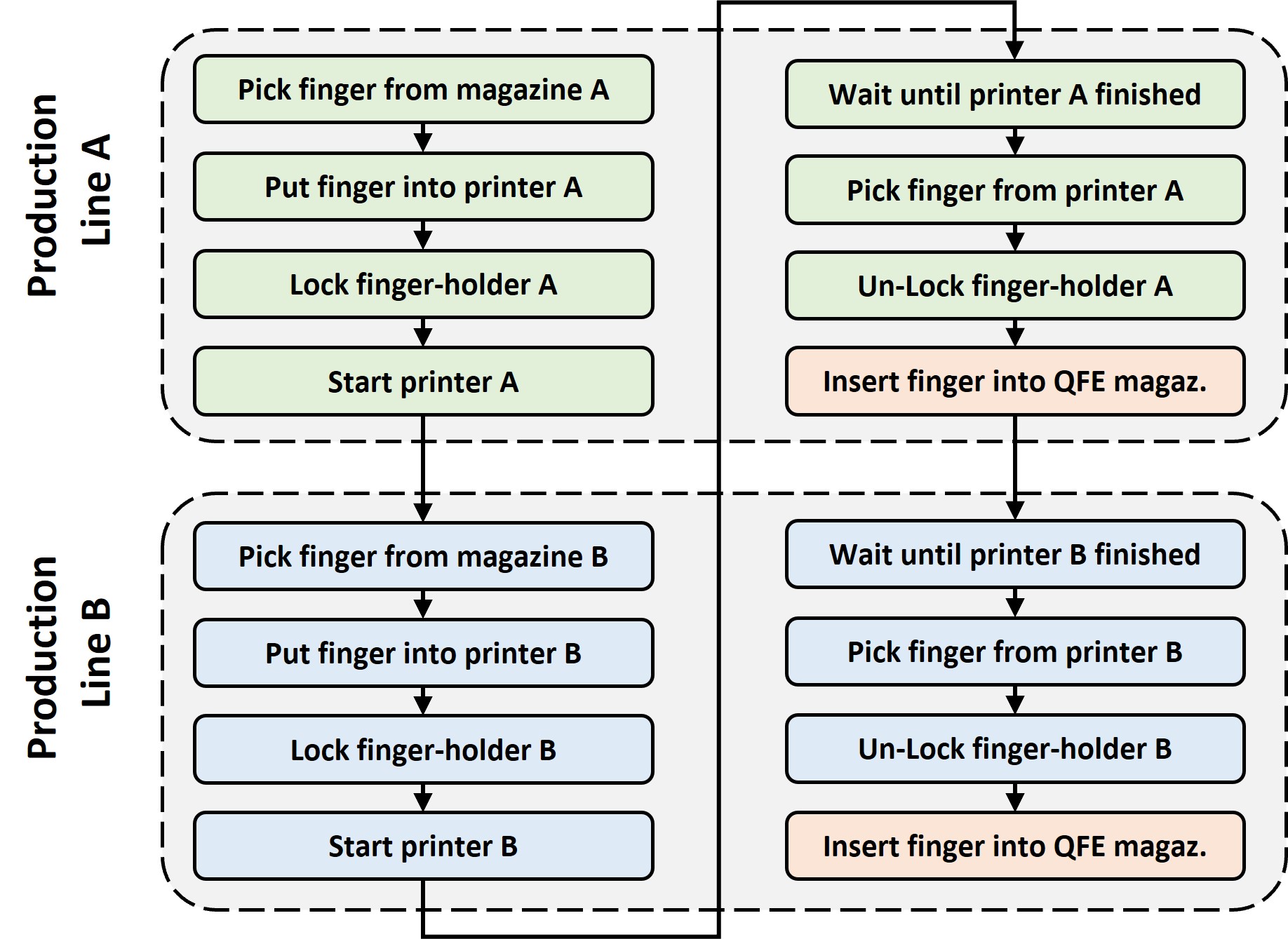}}
	\caption{A flowchart of the fingertip production cycle, described in Section \ref{pu}.}
	\label{fig:Fingertip-Production-Process}
	\vspace{-0.1cm}
\end{figure}
%
% PRODUCTION-UNIT
%-------------------------------------------------------------
\subsection{Production Unit}\label{pu}
In order to produce a customized finger-pair, the following steps must be conducted (see Fig.\ref{fig:Production/Task-Units} and Fig.\ref{fig:Fingertip-Production-Process} ). First, robot-A picks a finger-base from the finger-base\footnote{The finger-bases have been pre-produced by 3D printing.} storage magazine-A, inserts it into the finger-holder of printer-A and locks the finger-holder to prevent uplift movements of the inserted finger-base. This enables the 3D printing-pipeline to print a pre-defined fingertip on top of the inner finger-base surface\footnote{The used material for finger-base and fingertip is PLA.} (see Fig.\ref{fig:Production/Task-Units}). After the printing process is finished, the finger is picked up again by robot-A and inserted into a designated quick-finger-exchange magazine. After the printing process of the first finger is started at printer-A, the production process of the second finger is triggered in parallel (see Fig.\ref{fig:Fingertip-Production-Process}). Accordingly, robot-A picks another finger-base from finger-base magazine-B, inserts it into printer-B and places the finger next to the first finger into the designated quick-exchange finger magazine after the printing process is finished. 
The customized finger-pair can now be taken from the quick-finger-exchange magazine by robot-B, to conduct the designated task.

\begin{figure*}[ht!]
	\centerline{\includegraphics[width=17cm]{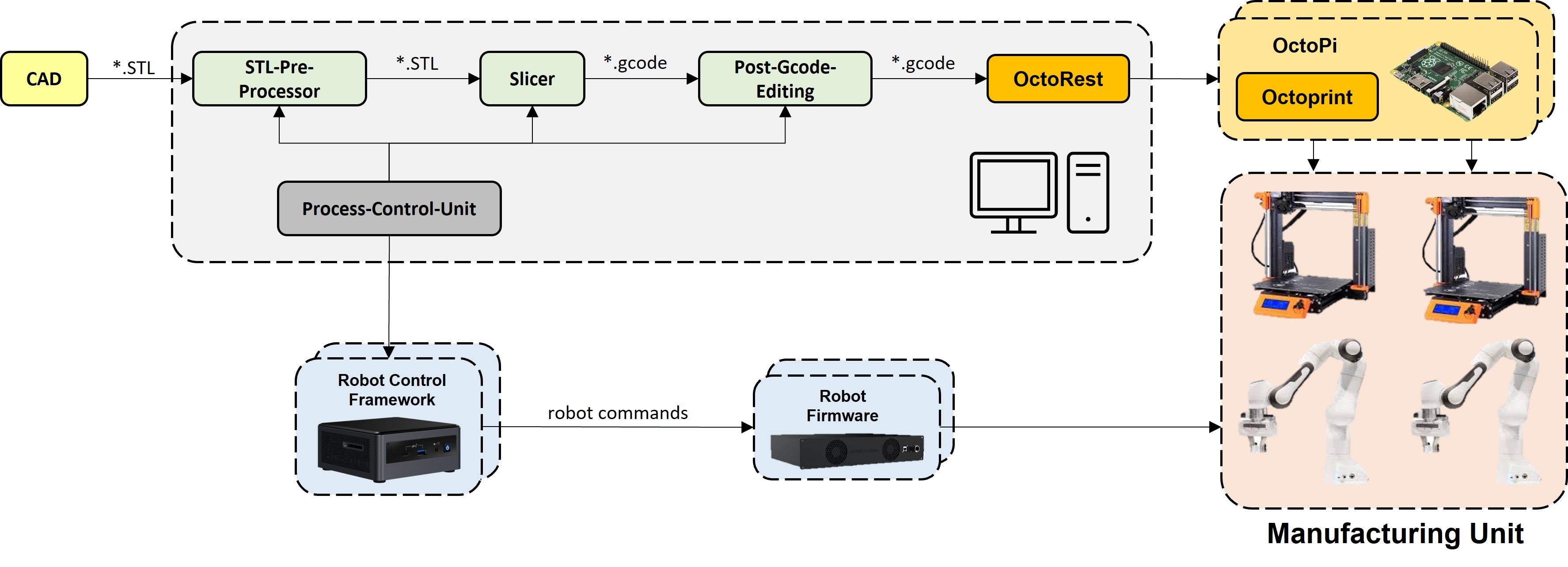}}
	\vspace{-0.2cm}
	\caption{
    Software pipeline running on five different device types: a device control PC, two robot-control computers, and master-controllers, as well as two raspberry pi's and 3D printers. The device control PC coordinates all production and task-execution steps.
    The fingertip CAD model is converted to STL-format and then pre-processed to align the fingertips with the finger-holder on the 3D printer build-plate. A slicer generates the gcode based on these processed STL files, which is additionally modified to avoid collisions between print-head and finger-holder. The processed gcode-file is uploaded to a raspberry pi 4b running OcotPrint via the python OctoPrint API OctoRest. The Raspberry pi's together with OctoPrint are enabling the communication between the process control PC and the 3D printers.
	% Robot-Arms
	The two panda robot-arms used to conduct the production and task-execution operations are controlled by a customized robot-control-framework and the firmware running on the panda master controller. The process control unit is calling the functionality from this robot-control-framework.}
	\label{fig:Software-Structure}
\end{figure*}

\begin{figure}[ht!]
	\centerline{\includegraphics[trim=0cm 0.5cm 0cm 0cm, clip, width=9cm]{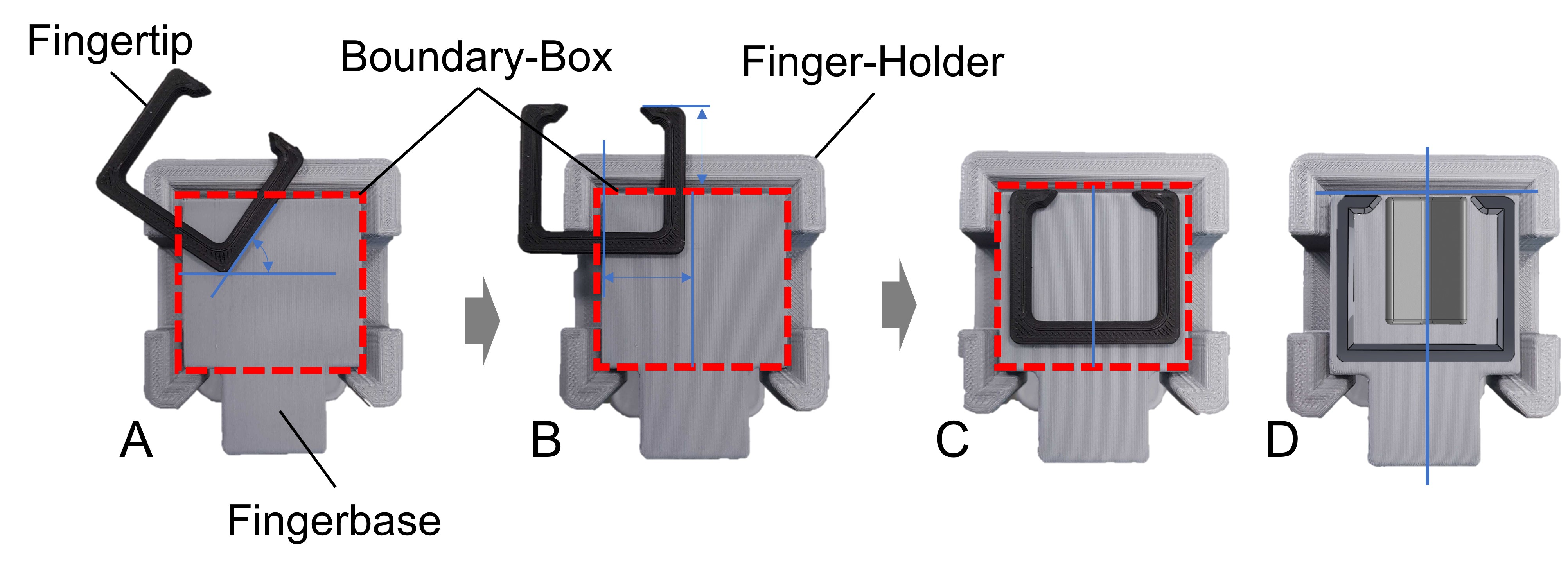}}
	\vspace{-0.1cm}
	\caption{
	STL pre-processing pipeline to re-orient (A), re-position (B), and center the fingertip relative to a virtual boundary box (C)/(D).} 
	\label{fig:STL-Pre-Processing}
	\vspace{-0.5cm}
\end{figure}

% Fingertip Design Guidelines
The fingertips of these fingers are designed manually, mostly based on the design guidelines of Greg Causey for gripper-finger geometries \cite{causey2003guidelines}. Accordingly, we mimiced the contour of the manipulation objects to maximize form-closure and closure force distribution. But contrary to Causey's recommendations we incorporated additional V shape chamfers for the key and battery fingertips to better compensate potential positioning errors.\\
\\
Performing the previously mentioned production cycle requires six software modules, illustrated in Fig.\ref{fig:Software-Structure} and described as follows.
\paragraph{Process Coordination}
The process control unit runs on a central host computer and integrates all the software modules and communication interfaces of the used devices. The communication between 3D printers and robot arms is conducted via Ethernet within a local network. Based on this setup, the process control unit controls and coordinates all the production and task execution steps. 
% Picture: ...
\paragraph{STL Pre-Processing}
After converting the CAD model of the custom-designed fingertip to an STL format (Standard Triangle Language), an STL pre-processing unit re-orients and automatically re-positions the model design onto the top of the finger-base (see Fig.\ref{fig:STL-Pre-Processing}). Each custom-designed fingertip might have a different global orientation and position relative to the printer coordinate system and correspondingly the finger holder. To place them at the desired position, two transformation steps are performed. First, the required re-orientations are performed manually\footnote{Note that this manual re-orientation is not a part of our automatic process, rather it should be done during the design of the CAD model.} once for each fingertip-model. 
To do so, the original position $(v_x, v_y, v_z)^T$ of the model is re-oriented to the new position $(v_{x_{rotate}}, v_{y_{rotate}}, v_{z_{rotate}})^T$ by multiplying it by a rotation matrix\cite{bajd2013introduction}.

Then the boundary box of the rotated model can be acquired by traversing all the vertices, represented with $v_{x_{min}}, v_{x_{max}}, v_{y_{min}}, v_{y_{max}}, v_{z_{min}}, v_{z_{max}}$ indicating the minimum and maximum size boundaries. These boundaries are compared with a virtual boundary box of size $b_x, b_y$ representing the finger-base print-area (see Fig.\ref{fig:STL-Pre-Processing}). The position of each vertex after translation is calculated as follows:
\begin{align*}
v_{x_{trans}} &= v_{x_{rotate}} - (v_{x_{max}} - v_{x_{min}})/2 - v_{x_{min}}\\
v_{y_{trans}} &= v_{y_{rotate}} - v_{y_{max}} + b_y/2 \\ 
v_{z_{trans}} &= v_{z_{rotate}} - v_{z_{min}}.
\end{align*}

The virtual boundary box itself is then manually positioned relative to the build-plate. This step has to be conducted only one time for each printer. All custom fingertips which fit into the boundary-box will be automatically centered and aligned to the front-edge of the finger-base as described.

The global z-offset is conducted via the slicer parameter to adapt the distance between the nozzle and finger-base independent of the STL-processing.
% Picture: ...

\paragraph{Automatic Slicing}
After the STL-files have been pre-processed, we use PrusaSlicer to generate the gcode needed to perform the printing process \cite{PrusaSlicerWeb}. PrusaSlicer provides command-line utility which can be interfaced with our process control unit. Important parameters like infill-density, layer-height, build-plate adhesion options or post-print-commands are configured via this interface.
\paragraph{Gcode Editing}
The generated gcode is adapted to avoid a possible collision between the finger-holder and print-head during the homing operations at the beginning of each print. Additionally the mesh bed leveling operations needed to be deactivated, via this process, to avoid collisions.
\paragraph{OctoPrint}
Once the gcode is ready, it can be uploaded to the OctoPrint-Server via the python OctoRest API \cite{OctoRestWeb}. OctoPrint \cite{OctoPrintWeb} is a browser based remote control and monitoring software for desktop 3D-printing applications. The software is installed and running via OctoPi on a raspberry pi 4b - enabling access to the printer via the local Ethernet network. In this setup, the process control unit can upload the generated gcode, give the order to print the fingertip, and monitor the state of the printing process. 
\paragraph{Robot-arm Operation Control}
The process control unit commands all required robot pick and place operations of the production as well as the task execution unit. It uses a custom Python/C++ framework. 
The framework was originally developed to enable complex manipulation task learning \cite{johannsmeier2019framework, Voigt2021a}. It provides different skills like move to contact or object insertion to conduct these robot operations via a python interface. It is executed on an Intel NUC computer, which interfaces the master-controller of the used panda robots with the local network. 

% % Picture: ...
% \begin{figure*}[ht!]
% 	\centerline{\includegraphics[width=17cm]{Figures/QEM-Working-Principle_v2-01.jpg}}
% 	\vspace{-0.3cm}
% 	\caption{Finger change process: (1) Gripper positioning above the trigger-tongues of the QFE magazine, (2) Establishing contact and force application to the QFE magazine. (3) Trigger-tongues of the QFE magazine push the spring pre-tensioned secure stone up. Then the gripper-finger-interfaces move to the center until the QFE-tongue of the finger-bases is fully inserted into the QFE-mechanism. (4) The gripper moves up and releases the secure-stone and locks the fingers-bases.}
% 	\label{fig:QEM-Mechanism}
% 	\vspace{-0.3cm}
% \end{figure*}
%
% TASK-EXECUTION-UNIT
%------------------------------------------------------
\subsection{Task-Execution-Unit}
The task execution unit contains another panda robot arm, and is equipped with a quick-finger-exchange mechanism, three quick-exchange magazines, an IoT-Box \cite{so2022towards} and a grasp-stability-test setup. The IoT-Box serves as a test station to evaluate the performance of the pick and insertion tasks with three different objects: a key, an Ethernet cable, and a battery. Different pick and insertion strategies are required to successfully conduct the tasks for these objects, enabling a more extensive evaluation of the presented approach. 
% Picture: ...
\begin{figure*}[ht!]
	\centerline{\includegraphics[width=17cm]{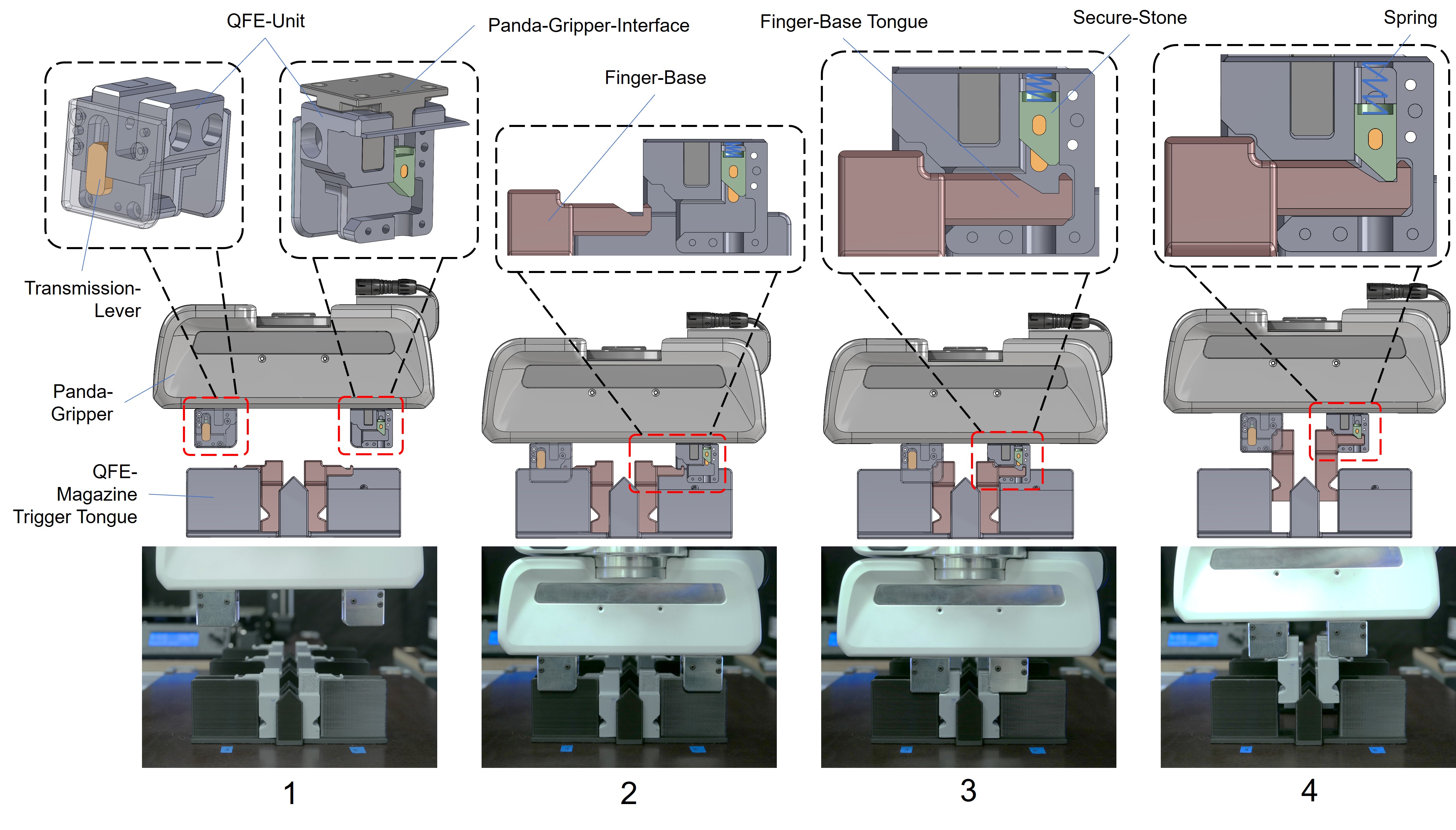}}
	\vspace{-0.3cm}
	\caption{Finger change process: (1) Gripper positioning above the trigger-tongues of the QFE magazine, (2) Establishing contact and force application to the QFE magazine. (3) Trigger-tongues of the QFE magazine push the spring pre-tensioned secure stone up. Then the gripper-finger-interfaces move to the center until the QFE-tongue of the finger-bases is fully inserted into the QFE-mechanism. (4) The gripper moves up and releases the secure-stone and locks the fingers-bases.}
	\label{fig:QEM-Mechanism}
	\vspace{-0.3cm}
\end{figure*}
\paragraph{Quick-Finger-Exchange-System}
The presented quick-finger-exchange (QFE) system enables the passive change of parallel gripper fingers during robot operations and is compact enough to be mounted to the finger-interfaces of a panda gripper. The mechanism uses a form-closure based locking-system - similar to a door-locking-mechanisms (see Fig.\ref{fig:QEM-Mechanism}). A secure stone holds the corresponding tongue of the finger-base in position.

The finger-bases are stored and positioned within the QFE magazines. In order to receive the fingers, the following process is conducted: (1) The QFE units of the gripper are positioned above the trigger tongues of the QFE magazine. (2) This allows the insertion of the trigger tongues into corresponding magazine tongue slits of the QFE elements. Once inserted, the gripper moves to contact and pushes the gripper towards the table plate. This pushes the secure stone of the QFE element up, since the transition levers transfer the QFE magazine tongue contact to the spring pre-tensioned secure stone. (3) Once the secure stone is pushed up, the gripper can close which inserts the finger-base QFE tongues into the corresponding opening of the QFE unit. (4) Are the finger-base tongues inserted, they can be secured by re-leaving the secure-stone via an upwards move of the gripper. A spring pushes the secure stone into the corresponding groove of the finger-base tongue and holds the finger via form-closure.  

% % Picture: ...
% \begin{figure*}[ht!]
% 	\centerline{\includegraphics[width=17cm]{Figures/QEM-Working-Principle_v2-01.jpg}}
% 	\vspace{-0.3cm}
% 	\caption{Finger change process: (1) Gripper positioning above the trigger-tongues of the QFE magazine, (2) Establishing contact and force application to the QFE magazine. (3) Trigger-tongues of the QFE magazine push the spring pre-tensioned secure stone up. Then the gripper-finger-interfaces move to the center until the QFE-tongue of the finger-bases is fully inserted into the QFE-mechanism. (4) The gripper moves up and releases the secure-stone and locks the fingers-bases.}
% 	\label{fig:QEM-Mechanism}
% 	\vspace{-0.3cm}
% \end{figure*}

	\section{Experiments}\label{exp}
The production as well as the task execution unit have been evaluated experimentally\footnote{The experiments have been documented in the accompanying video.}. 
% PRODUCTION EXPERIMENTS
%---------------------------------------------------------
\paragraph{Production Unit}
% Production Cycle Test
The production of the fingertips was evaluated by two kind of experiments: \\
(1) The complete production cycle was conducted for each custom-designed fingertip - starting with finger-base pick from the finger magazines and ending by inserting the task-specific finger-pairs into the QFE magazines. The printing times of each fingertip have been:
\begin{itemize}
    \item Key-fingertip: 5 min and 37 sec
    \item Ethernet-fingertip: 11 min and 16 sec
    \item Battery-fingertip: 9 min and 52 sec
\end{itemize}
These results show that a fast and fully automatic production of new gripper fingers is possible - potentially enabling a quick adaption of fingers to a new problem.\\
% Fingertip Geometry Measurements
(2) To evaluate the fingertip production consistency and robustness, the printing process of finger-tip \textit{A}\footnote{The  measurements of the other finger-types have not been analysed.} (manipulation object: key) was conducted 15 times with each printer - resulting overall in 30 printing trials. The geometrical dimensions of designated points of the finger have been measured via a caliper.
% Geometry Analysis and Explanations
Mean, variance and minimum/maximum values of the measurements for each point have been calculated (see \href{https://drive.google.com/file/d/1p7mTcgRQn_I6P4UWshcIvHr15j9dRT6k/view?usp=sharing}{Finger-Geometry-Data} ). The geometrical variation is significantly different for finger-base and finger-tip. The finger-base measurements show the smallest variance and absolute min/max differences of 0,15mm to 0,19mm. The finger-tips show a measurement variance and absolute value difference of 0,11mm-0,79mm which is many times higher compared to the finger-base. This is plausible, since the finger-base is hold in position within the finger-holder by form-closure during the printing process. This results in a small gap between finger-base and finger-holder sidewalls, giving room to potential small movements of the finger-base during the printing process. \\
Nevertheless, the analysis shows that the geometrical variations are small relative to the absolute finger dimensions. \\
% ??? REMOVE ??? --> Or shrink to footnote?
%-------------------------------------
Apart from this geometrical analysis, the first layer of all sample-prints show a slightly rough border, which seem not to influence the grasping performance. Also three specimens showed small 3D printing failures \footnote{Filament parts on the fingerbase body}. One specimen had this failure within the grasping area of the fingertip. 
%-------------------------------------
%
% % Picture: Finger-Measurements
% \begin{figure}[ht!]
% 	\centerline{\includegraphics[width=8.5cm]{Figures/Finger-Geometry-Measurement-Points_v1-01.jpg}}
% 	\caption{Measurement points taken from produced finger specimens.}
% 	\label{fig:Finger-Geometry-Measurement-Points}
% \end{figure}
%
% % Picture: Finger-Measurement-Analysis
% \begin{figure*}[ht!]
% 	\centerline{\includegraphics[width=17.5cm]{Figures/Finger-Geometry-Analysis_dummy_v1-01.jpg}}
% 	\caption{Finger geometry measurement analysis results. The table shows mean and variance of geometry measurements taken at designated points (see Fig. \ref{fig:Finger-Geometry-Measurement-Points}) from 15 finger specimens produced by each printer.}
% 	\label{fig:Finger-Geometry-Measurement-Analysis}
% \end{figure*}
%---------------------------------------------------------

% TASK EXPERIMENTS
%---------------------------------------------------------
\paragraph{Task Execution Unit}
The produced fingers and the presented QFE system are further evaluated by performing pick and insert operations using the described IoT-Box as well as a grasp stability test. Ten different trials have been conducted. One trial contains the following series of steps and experiments for each finger-type and corresponding manipulation object key, ethernet-cable and battery:
\begin{enumerate}
    \item Insert finger-pair into QFE mechanism
    \item Regular task-experiments
    \item Non-regular task-experiments
    \item Grasp-stability test
    \item Regular task-experiments with position offsets
    \item Place finger-pair back to QFE mechanism
\end{enumerate}
%
% QFE Pickup
First step is the insertion of a finger-pair into the QFE mechanism, in order to conduct the following task experiments.

\begin{figure*}[ht!]
	\centerline{\includegraphics[width=17cm]{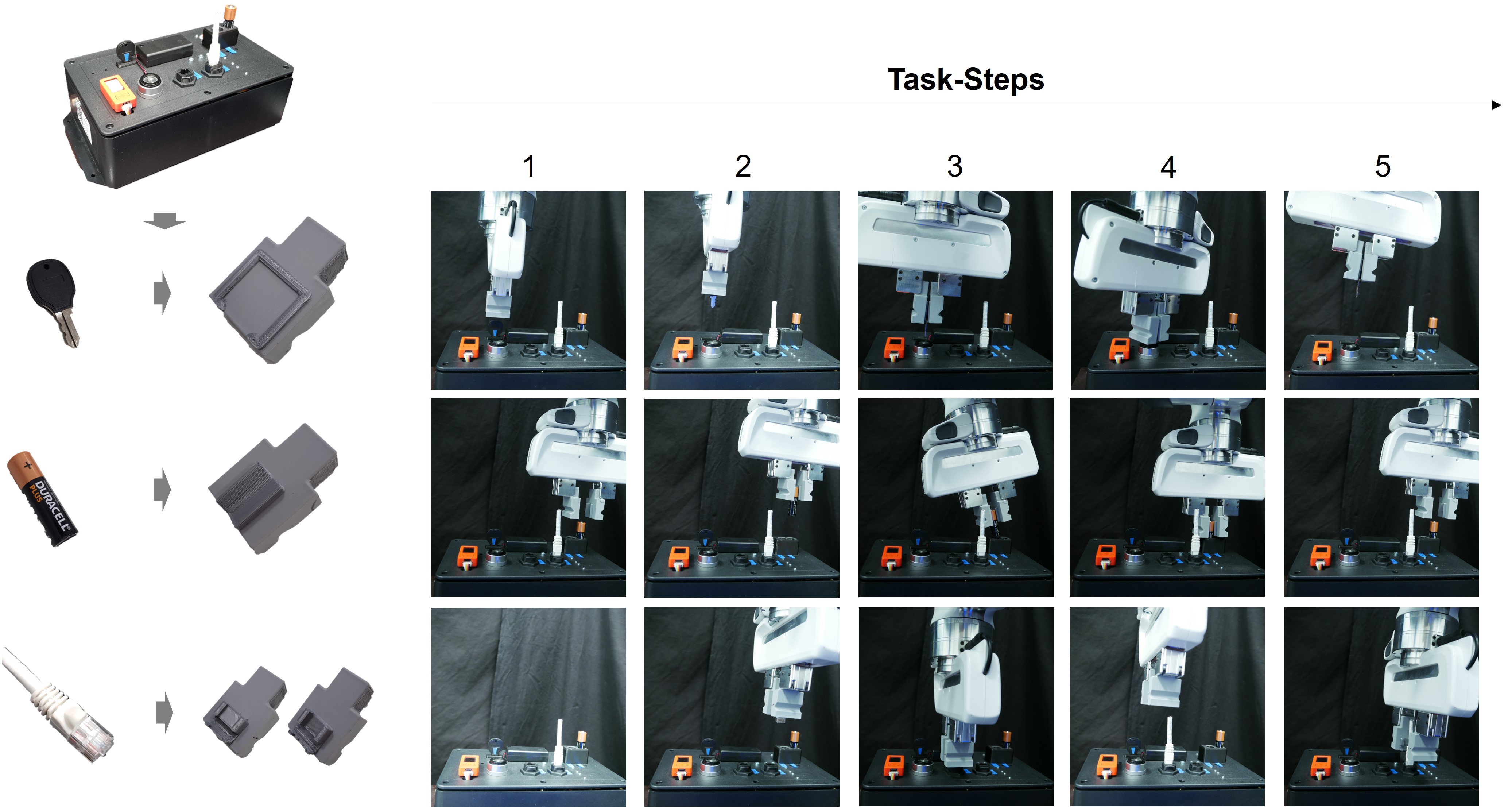}}
	\caption{Regular Task Experiments conducted via the IoT-Box: (task-A) Pick key, insert into lock and turn it. (task-B) Pick battery and insert into battery-holder. (task-C) Pick Ethernet-cable, and insert into corresponding Ethernet-plug. 
	%Therefore three different manually designed fingertips have been automatically produced and placed to the corresponding QFE magazines by the described pipeline. The robot-arm of the task-execution unit, conducted the described tasks after picking the required finger-pair from the QFE magazine.
	}
	\label{fig:Experiment-Steps}
	\vspace{-0.3cm}
\end{figure*}

% Regular task experiments
The \textbf{regular task-experiments} evaluate the task execution, the fingertips are actually designed for, with the following main steps (see Fig.\ref{fig:Experiment-Steps}): (1) Pick and extract the manipulation object from its original storage. (2) Insert the object into a new target slot. (3) Pick the object again and (4) insert it back to its original storage slot. In case of the key, an additional key turning step is conducted after inserting the key into its target-slot - the lock, to validate full insertion.
% Success-Definition
A task experiment is counted as successful if all previously mentioned steps have been successfully conducted.

% Fingertip Generalization Performance
To evaluate the \textbf{generalization} abilities of the designed and automatically produced fingertips, we conduct the task-experiments with the same finger pair also for the other manipulation objects (non-regular task-experiment). E.g., we execute the task-experiments for the ethernet-cable and battery scenario with the finger-pair designed for the key object.
This should give insights how universal the different fingertip prototypes can be used.

\begin{figure}[ht!]
	\centerline{\includegraphics[width=8.5cm]{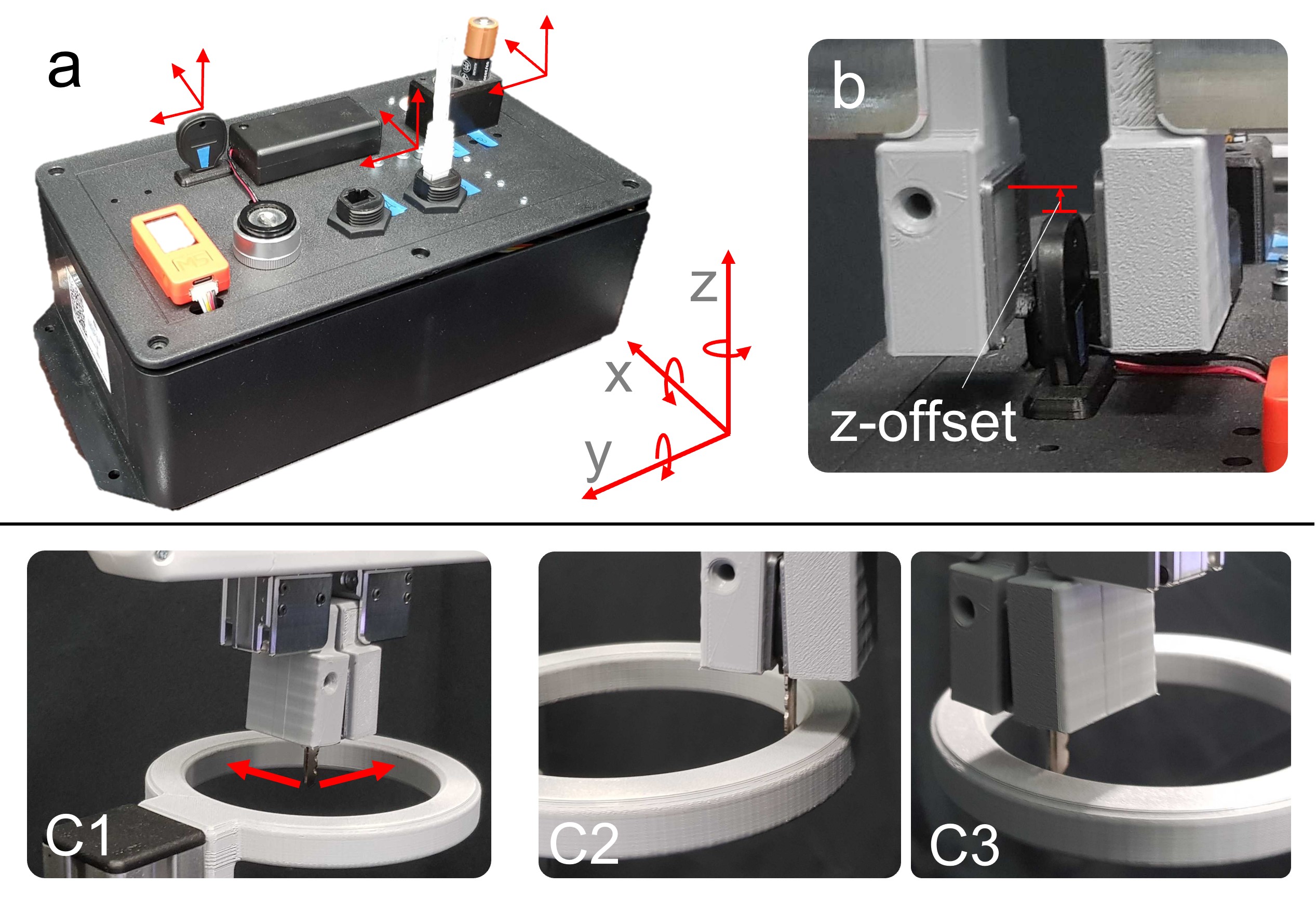}}
	\vspace{-0.2cm}
	\caption{(a) Coordinate axis indicating the used offset translation and rotation directions. (b) Z position offset visualization for key manipulation object example. (C1-3) Grasp stability test setup and visualization of applied pushing directions. The test evaluates if the end-effector moves beyond the ring-border to check for ineffective object grasping.}
	\label{fig:Offset-and-Grasp-Stability-Experiment-Description}
	\vspace{-0.3cm}
\end{figure}

% Grasp Stability Test
Further we evaluate the \textbf{grasp stability} by grasping the manipulation object and pressing it against the inner surface of a ring with 5N pushing force in two different directions (see Fig. \ref{fig:Offset-and-Grasp-Stability-Experiment-Description}). In order to check if the grasped object is loose and does move during the test, the cartesian position of the robots endeffector after the pushing execution is compared to the expected position before the pushing action. The mean square error between both positions is calculated and compared with a critical limit - indicating if the key moved significantly ($>$ 5mm).
%
% % Picture: Grasp-Stability-Experiments
% \begin{figure}[ht!]
% 	\centerline{\includegraphics[width=6cm]{Figures/Grasp-Stability-Experiment-Concept_v1-02.jpg}}
% 	\caption{Grasp stability test setup and visualization of applied pushing directions. The test evaluates if the end-effector moves beyond the ring-border to check for ineffective object grasping.}
% 	\label{fig:Grasp-Stability-Experiment-Setup}
% \end{figure}

% Offset Experiments
Finally, the regular task-experiments are conducted with a series of \textbf{position-offsets} applied for each grasping approach pose\footnote{The pose right before the gripper grasps the manipulation object}. For each coordinate axis x, y, z, five different position offsets 1-5mm and three different rotation offsets 5-15 degree are applied (see Fig. \ref{fig:Offset-and-Grasp-Stability-Experiment-Description}). This results in 24 overall offset experiments which enable the evaluation of the execution robustness of the fingers regarding potential robot-arm position-errors. Single axis offset experiments enable a systematic and clear connection analysis between the position error (direction and absolute value) and the resulting task execution performance. Contrary, the application of combined offset-errors along multiple axis applied to the same time, make a clear identification of the performance influencing parameters difficult if not impossible.
%
% % Picture: Offset-Experiments
% \begin{figure}[ht!]
% 	\centerline{\includegraphics[width=8.5cm]{Figures/Offset-Experiment-Concept_v1-02.jpg}}
% 	\caption{(a) Z position offset visualization for key manipulation object example. (b) Coordinate axis along which position and rotation offsets have been conducted.}
% 	\label{fig:Offset-Experiment-Description}
% 	\vspace{-0.3cm}
% \end{figure}
% Picture: Offset-Experiments & Grasp-Stability-Test-Setup

\begin{figure}[ht!]
	\centerline{\includegraphics[trim=0.5cm 0.4cm 0.7cm 0.75cm, clip, width=8.5cm]{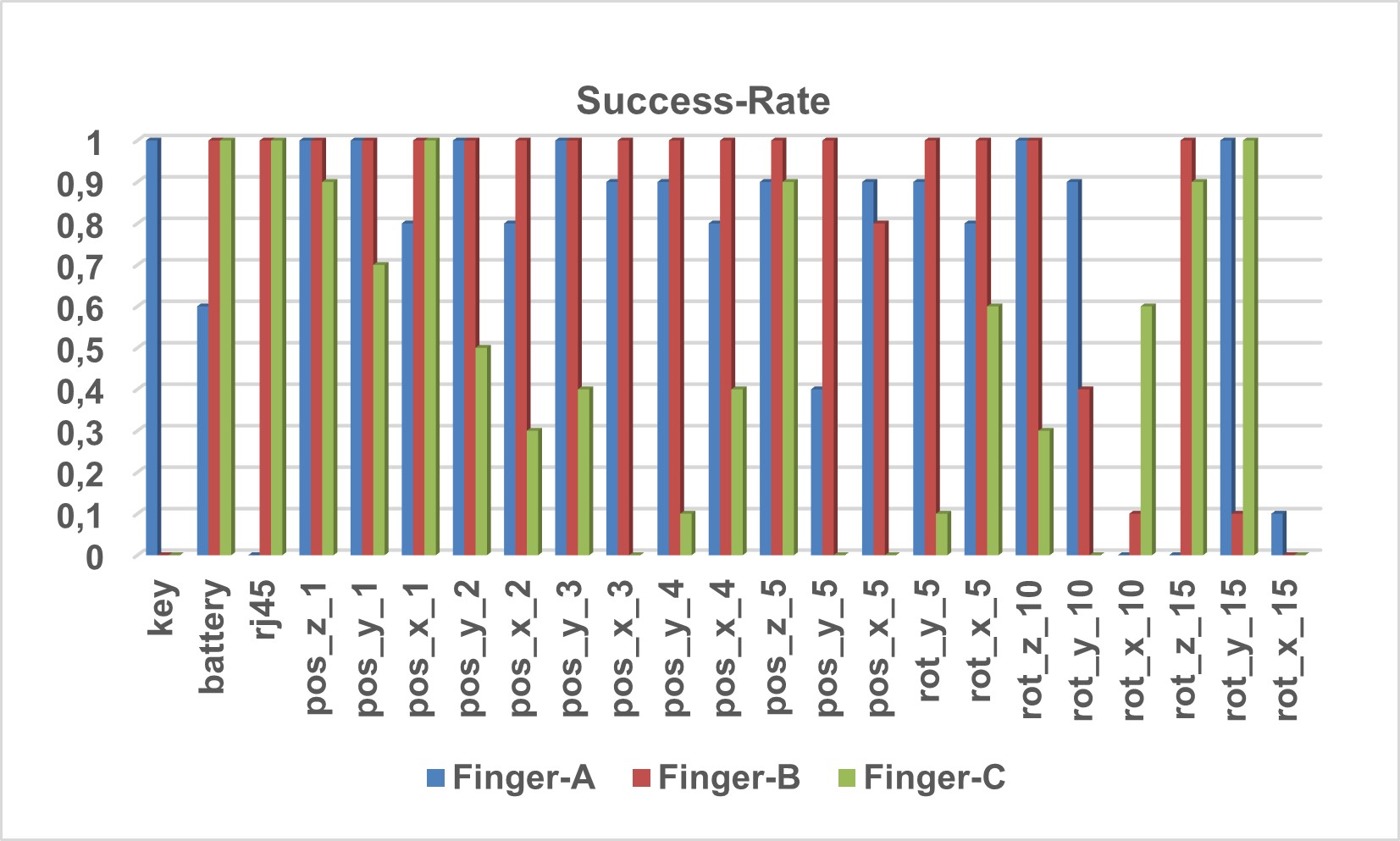}}
	\caption{Success rates of all task experiments (conducted by the three different finger-types) which show a success-rate unequal to one for at least one finger-type. The remaining task experiments not shown here, have a success-rate of one for all finger-types.}
	\label{fig:Task-Experiment-Success-Rates}
	\vspace{-0.3cm}
\end{figure}

% Result Description
\textbf{Results:}
For each manipulation object, 28 experiments have been conducted per trial. Accordingly, ten trials for three different manipulation objects result in 840 overall experiments. Fig.\ref{fig:Task-Experiment-Success-Rates} contains the success-rates for those experiments - conducted by a certain finger-type - where at least one experiment shows a success-rate smaller then one. The other not shown experiments showed all a success-rate of one. \newline
It can be concluded that the regular tasks, the lower offset error experiments and the grasp stability tests could be robustly executed for all manipulation-objects. On the other hand larger offsets caused different results for the tested fingers. While all finger-types could handle the z-position offsets robustly, the medium to large rotation offsets caused unreliable task execution of nearly all finger-types. Interestingly, the different finger-types show success-rates of 100\% for certain particularly large rotation offsets. These offsets in combination with the individual geometry seem to create a new stable grasping configuration, enabling a success-full task-conduction despite significant offset errors. Beside this general observations, the different finger-types showed the following individual characteristics:\\ 
% Finger-A: Key
Finger \textit{A} (key object), shows problems regarding the larger y-position, z-rotation and x-rotation offsets. The remaining position offsets could be handled with a relatively high success-rate.
% Finger-B: Battery
Finger \textit{B} (battery object), shows the highest robustness of all fingers. The biggest problems occured for the larger y and x rotation offsets.
% Finger-C: Ethernet-Cable
Finger \textit{C} (ethernet-cable), displays leaking robustness especially for the y-position as well as the y and x rotation offsets. \\
% Generalization
Regarding \textit{generalization}, finger \textit{B} (battery object) and finger \textit{C} performed best, while finger \textit{A} (key object) was not able to successfully execute the tasks for the other objects (battery and ethernet-cable). This can be explained by the significantly bigger V-chamfer shape of fingertip B compared to finger A and supports the assumption that V-shaped fingertips increase grasping-robustness. 
%
% Picture: Task-Experiment-Results (Success-Rates)
	\section{Discussion and Conclusion}\label{cond}
% Finger Production
The experimental results show that our approach successfully provides an improved level of finger production and usage automation. 
% Time/Cost Efficient Production Change
This level of automation enables a time and cost efficient change of the production setup for a new assembly task. Accordingly, our approach is a step further to fully customized one batch size product/system production.

% Reduced Finger Iteration Time
Additionally, the provided 3D printing pipeline potentially reduces the iteration and adaption time of novel gripper finger designs since multiple manual steps have been automated and the decoupling of finger-base and finger-tip production reduces the required printing time significantly, since only the smaller finger-tip must be printed ad hoc. 
This decoupling reduces the design complexity as well, due to the fact that only the fingertip geometry must be considered and defined.
%Furthermore, decoupling the finger-base from the fingertip decreases the design complexity since only the fingertip geometry must be considered.
%\addtext{Also the printing time reduces dramatically since only the much smaller fingertip must be printed. This further reduces the adaption time of a finger-gripper-setup to a new production case.}

% Quick-Exchange-Mechanism
Additionally, the presented quick-finger-exchange system increases operation flexibility and performance of robot-arms, since multiple - for a certain operation specialized fingers - can be exchanged during the task-execution. The implementation of a finger-exchange system compared to an end-effector change approach, enables the continues usage of existing and potentially expensive gripper systems. This simplifies scaling the number of changeable "tools" from an economical as well as design complexity point of view significantly. 
%
% More task executable
Such an increasing number of "on the fly" changeable task-specific fingertips could increase the number of tasks and applications a single robot can conduct. Additionally, task specialized fingertips potentially increase manipulation performance and robustness. This enables the execution of complex tasks, composed by sub-tasks which require specialized fingertips to be successfully conducted. Accordingly, a robot with a quick-finger-exchange system can potentially perform a significantly larger number of tasks better and more robust, then a general multipurpose gripper finger system.
%\addtext{Reason how exactly our approach is ment to be used and why it is better then classical assemebly approaches.}

% Automatic Design Pipeline
This improved flexibility and performance could be further enhanced by advancing the current setup to a fully automatic fingertip identification pipeline. This pipeline could automatically design, produce and apply fingertips for a given set of manipulation objects and task scenarios. Such an approach would reduce required finger development and production setup times significantly, while decoupling the design quality from the individual skills and experience of a single designer.

% Novel Production Paradigm
In this context, the presented setup requires many different tactile manipulation operations, only possible by force/torque sensorized compliant robot-arms. Accordingly, the described pipeline could potentially build the foundation for a novel production paradigm: Tactile 3D manufacturing. The concept of fully automated, adaptive 3D printing based system manufacturing and iteration could potentially be applied to other application scenarios like sensor or actuator development as well.

% Future work
% -- Different finger-bases --
Accordingly, future work will focus on further improving and extending the current production and task execution setup. The currently used finger-base version is limited to a certain group of rather small manipulation objects, which can only be grasped at the outer surface. To manipulate objects with other form factors, different finger-bases would be required. A next iteration could integrate such a wider spectrum of different finger-base geometries. 
% -- Automatic fingertip design --
Furthermore, a fully automatic fingertip designer will be developed and interfaced with the production unit in order to achieve a fully automatic identification of the fingertips, required for a given manipulation scenario.

% Picture: ...
	\section{Acknowledgements}

The authors also would like to thank Robin Kirschner for her fruitful discussions and contributions. 
	
	% References:
	\bibliographystyle{IEEEtran}
	\renewcommand{\baselinestretch}{0.96}\normalsize % 0.97 in RAL paper
	\bibliography{references1}{}
	%\bibliography
    %{
	    %References/Regular-Hand-Design/ref_regular-hand-design,%
	    %References/Robot-Optimization/ref_robot-opt-gripper-finger,%
	    %References/Robot-Tool-Changer/ref_cnc-tool-changer,%
	   % References/Robot-Tool-Changer/ref_robot-hand_modular-fingers,%
	   % References/Robot-Tool-Changer/ref_robot-tool-changer,%
	   % References/Robot-Tool-Changer/ref_other-tool-changer,%
	   % References/3D-Printing/ref_3D-printing_slicer%
	%}{}
	
\end{document}